\documentclass[letterpaper]{article} 
\usepackage{aaai25}  
\usepackage{times}  
\usepackage{helvet}  
\usepackage{courier}  
\usepackage[hyphens]{url}  
\usepackage{graphicx} 
\usepackage{natbib}  
\usepackage{caption} 
\usepackage{algorithm}
\usepackage{algorithmic}
\usepackage{booktabs}
\usepackage{subcaption}

\usepackage{newfloat}
\usepackage{listings}
\DeclareCaptionStyle{ruled}{labelfont=normalfont,labelsep=colon,strut=off} 
\floatstyle{ruled}
\newfloat{listing}{tb}{lst}{}
\floatname{listing}{Listing}
\title{Synthetic Data Generation with LLM for Improved Depression Prediction}
\author{
    Andrea Kang \textsuperscript{\rm 1},
    Jun Yu Chen \textsuperscript{\rm 1},
    Zoe Lee-Youngzie \textsuperscript{\rm 1},
    Shuhao Fu \textsuperscript{\rm 1}
}
\affiliations{
    \textsuperscript{\rm 1}University of California, Los Angeles \\
    adkang@g.ucla.edu, jochen030327@g.ucla.edu, youngzielee@ucla.edu, fushuhao@g.ucla.edu
}

\usepackage{bibentry}

\begin{document}

\maketitle

\begin{abstract}
Automatic detection of depression is a rapidly growing field of research at the intersection of psychology and machine learning. However, with its exponential interest comes a growing concern for data privacy and scarcity due to the sensitivity of such a topic. In this paper, we propose a pipeline for Large Language Models (LLMs) to generate synthetic data to improve the performance of depression prediction models. Starting from unstructured, naturalistic text data from recorded transcripts of clinical interviews, we utilize an open-source LLM to generate synthetic data through chain-of-thought prompting. This pipeline involves two key steps: the first step is the generation of the synopsis and sentiment analysis based on the original transcript and depression score, while the second is the generation of the synthetic synopsis/sentiment analysis based on the summaries generated in the first step and a new depression score. Not only was the synthetic data satisfactory in terms of fidelity and privacy-preserving metrics, it also balanced the distribution of severity in the training dataset, thereby significantly enhancing the model's capability in predicting the intensity of the patient's depression. By leveraging LLMs to generate synthetic data that can be augmented to limited and imbalanced real-world datasets, we demonstrate a novel approach to addressing data scarcity and privacy concerns commonly faced in automatic depression detection, all while maintaining the statistical integrity of the original dataset. This approach offers a robust framework for future mental health research and applications. 
\end{abstract}

\begin{links}
    \link{Code}{https://anonymous.4open.science/r/LLM4Depression-C0A6}
\end{links}

\section{Introduction}

Over 300 million people worldwide suffer from major depressive disorder (MDD), a psychiatric disorder that is known to be the leading cause of disability and premature mortality \cite{chodavadia2023prevalence}. In particular, a number of studies have reported a substantial rise in the prevalence of depression disorders since the COVID-19 pandemic \cite{santomauro2021global}. Practitioners reported a shortage in capacity to meet this surge in demand for psychiatric care, however, leading to increased workloads for psychologists and longer waiting lists for patients \cite{apa_covid19_practitioner_2021}.

A state-of-the-art approach for handling various dialogues, including mental health assessments, is the usage of Large Language Models (LLMs) \cite{farruque2024depression}. These models have garnered significant attention due to their remarkable performance across a variety of natural language processing tasks. Since the release of OpenAI's ChatGPT \cite{gpt3} in 2022, LLMs have demonstrated substantial advancements in general-purpose language understanding and generation. They have evolved rapidly through training on extensive data and parameters, showcasing capabilities that extend beyond traditional NLP applications to tasks involving complex reasoning and interaction.

Despite the aforementioned potential, applications of LLMs and other ML methods in mental healthcare face unique challenges. First, the diagnosis of psychiatric disorders typically requires clinicians to observe and interpret patients' behaviors over multiple psychotherapy sessions. This is a complex decision-making process that integrates information about the patient's psychological, social, and environmental factors. This makes it challenging for artificial classifiers to accurately detect depressive symptoms from isolated health records. Second, there is limited availability of annotated health records for patients with a certain psychiatric disorder, and existing datasets are often small and imbalanced in terms of demographics or symptom severity. Because the performance of ML methods highly depend on the training data, a model trained on such skewed datasets can amplify their biases and may result in misleading or non-generalizable results. Finally, there is an obvious privacy concern as mental health data inherently contains highly sensitive information about the patient's background, mental health status, and personal experiences.

Synthetic data is often used as a solution for such challenges through augmenting data and introducing data privacy through anonymization, separating itself from any sensitive information tied to any individual or group.
While it is most commonly incorporated in computer vision \cite{borkman2021unity} \cite{man2022review}, there is growing interest in utilizing such data in critical areas such as medical \cite{kokosi2022synthetic} and financial \cite{assefa2020generating} fields.

In this paper, we propose and explore a novel pipeline for detecting depression from text transcripts of clinical interviews, augmented with synthetic data using an LLM.

\begin{figure*}[htbp]
\centering
\includegraphics[width=0.95\textwidth]{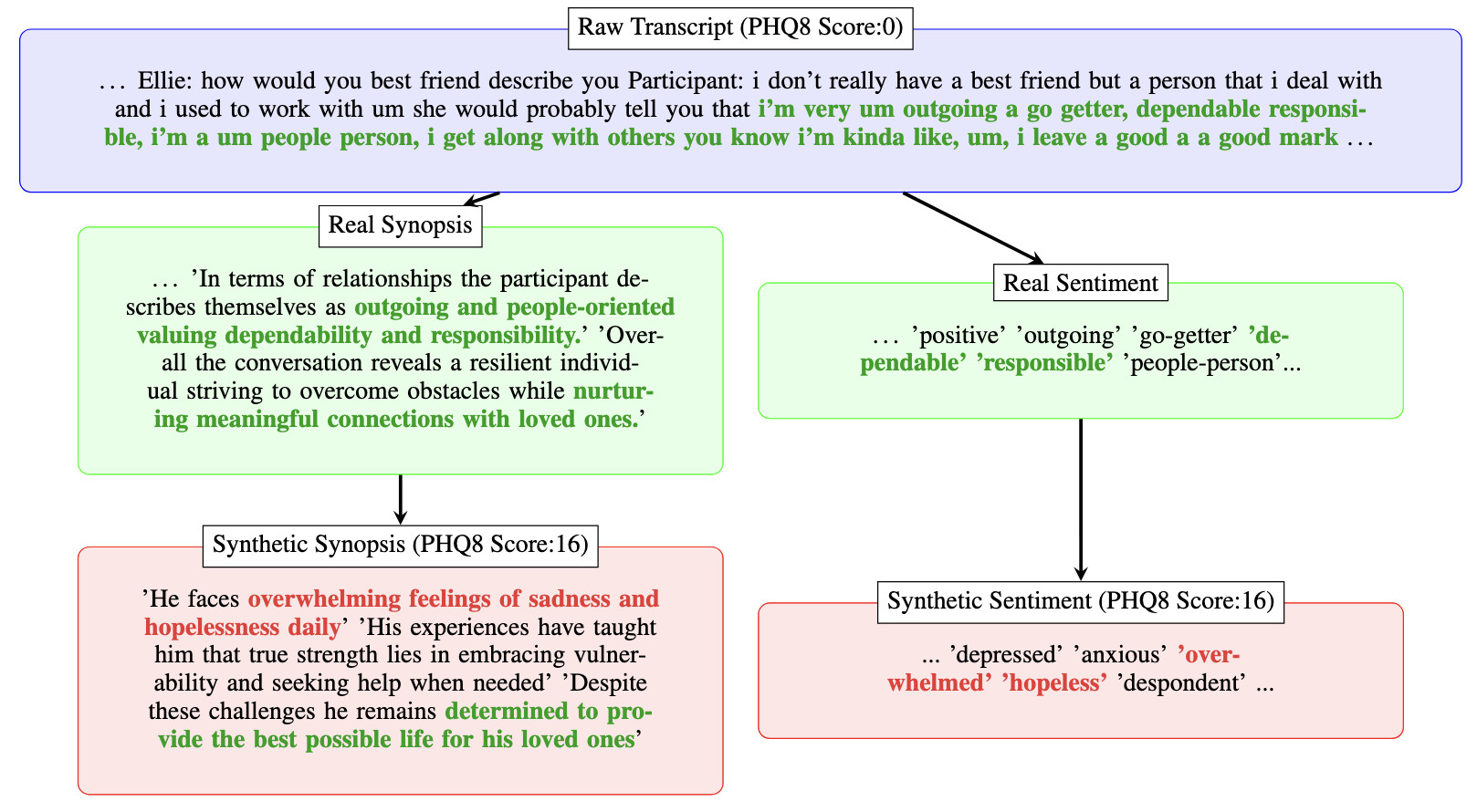}
\caption{A flowchart of the Chain-of-Thought pipeline. The LLM was able to capture the transcript's key moments in both the synopsis and sentiment analysis. Once provided the higher PHQ-8 score, the LLM could generate a new synopsis and sentiment analysis that maintain the original details (in green) while showcasing a more depressed character (in red).}
\label{fig:cotpipe}
\end{figure*}

\section{Related Work}


The potential of using advanced machine learning methods as useful tools in psychiatric healthcare has been demonstrated in recent years \cite{thieme2020machine, iyortsuun2023review}. For instance, machine learning methods have been used to predict the presence of major depressive disorder (MDD) using biomedical and demographic data from electronic health records \cite{nemesure2021predictive}. Similarly, unstructured text data from online social networks (OSNs) can be analyzed to identify patterns associated with MDD \cite{abd2020application, chiong2021textual}. Such approaches enable earlier detection of symptoms, which is critical for minimizing the adverse effects of MDD \cite{picardi2016randomised}. Additionally, these techniques facilitate more personalized prognoses while reducing the workload of mental health practitioners.

In particular, synthetic data generation using generative models are now increasingly employed to produce data that resembles real-world data while ensuring patient privacy.  For instance, Generative Adversarial Networks (GANs) are used to generate time-series data \cite{dash2020medical} and medical images \cite{shin2018medical}, while language models are utilized for generating synthetic health records \cite{yuan2024continued} and responding to healthcare-related queries \cite{med_palm}.

\subsection{Machine Learning for Depression Detection}

While there are multiple metrics to diagnose and determine the severity of a patient's despressive disorders, the most commonly used metric is the Patient Health Questionnaire-8 (PHQ-8) score \cite{kroenke2009phq}.
This eight-item self-report questionnaire asks patients to evaluate the intensity of their depressive symptoms over the past two weeks. For each symptom e.g., "Little interest or pleasure in doing things", the patient would answer on a scale of 0 to 3, where 0 indicates "Not at all" and 3 indicates "Nearly every day". The total score can range from 0 to 24, where a score of 10 is the cutoff to define depression.

Detecting depression-related symptoms from textual as well as visual and auditory data has been a popular area of research. With advances in deep and machine learning methods, studies have utilized convolutional neural networks (CNNs) or recurrent neural networks (RNNs) to predict risk of depression from text data, ranging from social media posts \cite{orabi2018deep} \cite{squires2023deep} to transcripted clinical interviews \cite{trotzek2018utilizing}. Additionally, multi-modal combination of visual and auditory features observed during the interviews can be used to capture nuances that are not included in the text transcripts \cite{muzammel2021end}. One particular study uses a Random Forest estimator to predict PHQ-8 scores from visual, auditory, and text features from DAIC \cite{rf} but the performance of this algorithm was best when using text features extracted from the transcripts.

More recent advances have led to a surge of research projects that use LLMs for this endeavor. For example, LLMs such as GPT-3.5 \cite{gpt35}, GPT-4 \cite{gpt4o}, BERT \cite{devlin2018bert}, and LLaMA \cite{Touvron2023LLaMAOA} have been utilized to detect symptoms of depression from social media data \cite{wang2024explainable, bao2024explainable, farruque2024depression}
or predict scores on the PHQ-8 from clinical interview data \cite{ohse2024zero, danner2023advancing, sadeghi2023exploring, hadzic2024enhancing}. In particular, the Dual Encoder model \cite{dual_encoder} has achieved strong performances through the utilization of prefix-based adaptation alongside the general-purpose embedding to tune the language model for depression detection.

\subsection{Synthetic Data Generation in Healthcare} 

Recent studies demonstrate that limited data availability, particularly due to privacy concerns, can be overcome by generating synthetic data. \cite{murtaza2023synthetic}
Synthetic data generation can also diversify the dataset further by incorporating a variety of linguistic styles, thereby making the LLM robust to language variation. This has led to improvements in detecting critical depressive features, such as suicidal ideations from social media posts \cite{ghanadian2024socially} to difficult-to-treat depression from sensitive medical records \cite{lorge2024detecting}.

Beyond general-purpose LLMs, specialized models have been developed with synthetic data to address domain-specific problems. For example, Asclepius \cite{kweon2023publicly} and HEAL \cite{yuan2024continued} are LLMs trained on synthetic clinical records that maintain their accuracy on real-life data, providing high-quality responses to healthcare-related queries. 

\begin{figure*}[ht]
    \centering
    \begin{subfigure}[b]{0.45\textwidth}
        \centering
        \includegraphics[width=\textwidth]{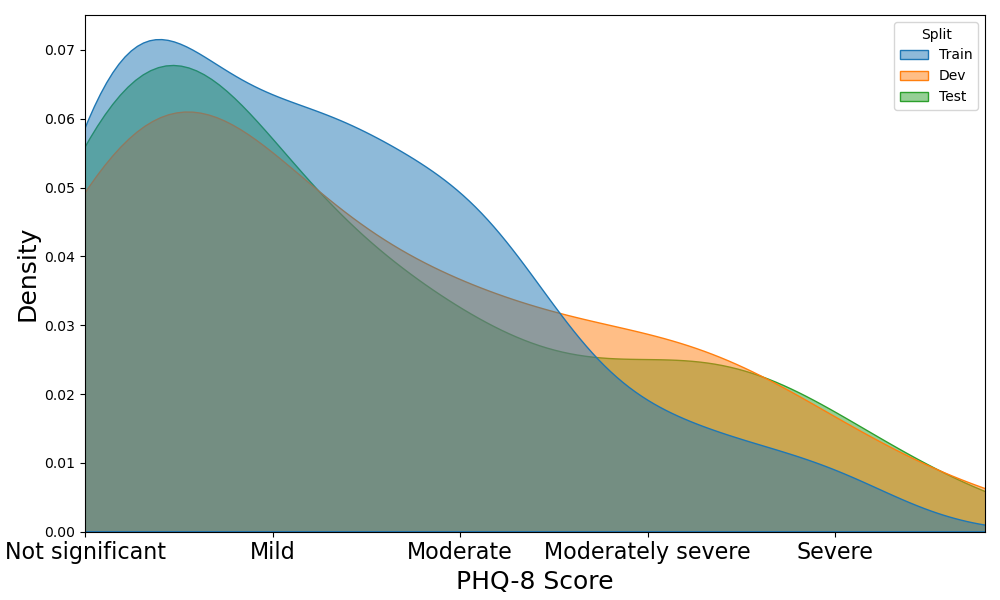}
        \caption{Original Data Distribution}
        \label{fig:original_distribution}
    \end{subfigure}
    \hfill
    \begin{subfigure}[b]{0.45\textwidth}
        \centering
        \includegraphics[width=\textwidth]{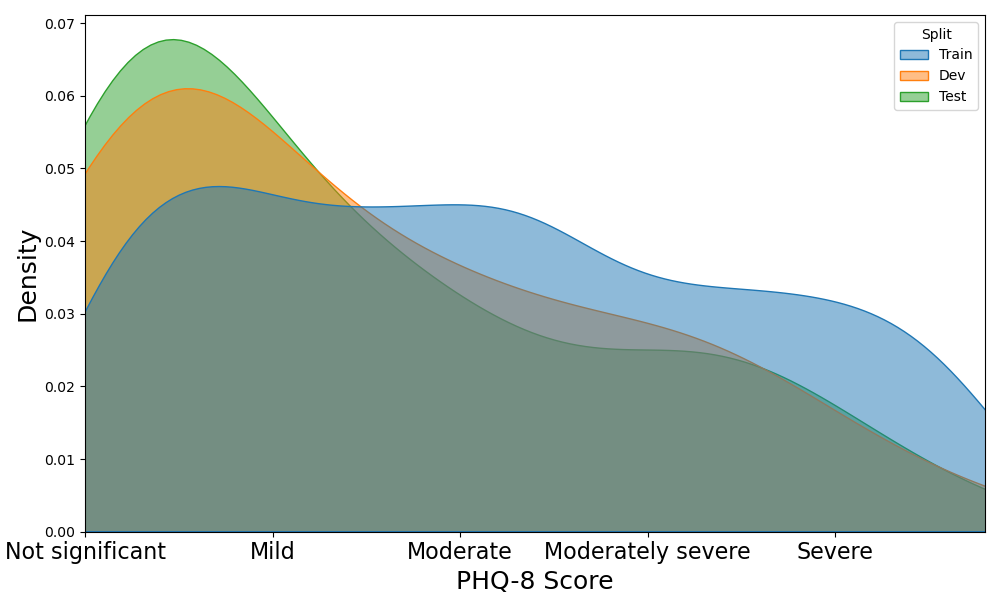}
        \caption{Original + Synthetic Data Distribution}
        \label{fig:synthetic_distribution}
    \end{subfigure}
    \caption{Data distribution for different PHQ-8 scores in the original (left) and combined datasets (right).}
    \label{fig:data_distributions}
\end{figure*}

In this paper, we propose a pipeline for generating synthetic data through chain-of-thought prompting in large language models (LLMs). The proposed pipeline addresses the aforementioned challenges in data scarcity and privacy by creating a larger and more balanced synthetic dataset derived from clinical interview transcripts. To evaluate the effectiveness of this approach, the synthetic dataset is assessed based on three key criteria: fidelity, examining whether the synthetic data aligns with the distribution of the original data; utility, evaluating improvements in depression severity detection; and privacy, assessing whether the synthetic data effectively protects participants from identification. This work leverages the exceptional generalizability of LLMs trained on large-scale language corpora\cite{ohse2024zero}.

\section{Synthetic Data Generation}

\subsection{Dataset}
The Distress Analysis Interview Corpus-Wizard of Oz (DAIC-WOZ) dataset \cite{daic} is utilized for detecting Major Depressive Disorder (MDD) through multi-modal analysis of interviews. It includes audio, video, and text data from 189 subjects, each engaged in interviews conducted by an animated virtual interviewer named Ellie. The dataset features interviews ranging from 7 to 33 minutes, with audio features extracted via the COVAREP toolkit, video features using the OpenFace toolkit, and semantic features analyzed with the Linguistic Inquiry and Word Count (LIWC) software.

In this project, we utilize the text data from the DAIC-WOZ dataset, which consists of transcripts of interviews between participants and a virtual interviewer, Ellie. The transcripts are verbatim responses of the participants to Ellie's questions, covering a wide range of topics related to personal experiences, emotions, and behaviors.

The DAIC-WOZ dataset exhibits a notable imbalance - only 46 out of 189 patients scored 10 or higher on the PHQ-8, indicating that the majority of patients were not diagnosed with depression. To address this imbalance and ensure a more equitable representation of higher PHQ-8 scores in our training data, we employ an over-sampling technique for datapoints with elevated PHQ-8 scores during synthetic data generation, as detailed in the following sections. Figure \ref{fig:original_distribution} illustrates the distribution of PHQ-8 scores across the training, development, and test datasets.

\subsection{Reconstructing Transcripts into Synopsis and Sentiment Analysis}
Through our experiments, we identified two key issues with directly generating full transcripts using Llama 3.2. First, the generated transcripts often include irrelevant elements such as breaks, stuttering, repetition, and off-topic conversations, which negatively impact predictive performance in downstream tasks. Second, the generated transcripts closely mimic the original data, raising concerns about potential privacy leakage. To address these challenges, we propose a two-step approach: first, generating a synopsis and sentiment analysis based on the original transcript (illustrated in Figure \ref{fig:cotpipe}), and second, generating a variant of the synopsis and sentiment analysis with a different depression severity level. The synopsis provides an objective summary of the transcript, capturing the patient's personal experiences and challenges, while the sentiment analysis highlights emotional aspects, including the type and intensity of emotions expressed. This method replaces verbose transcripts with concise summaries, offering clearer diagnostic insights into the patient's emotional state while mitigating privacy concerns.

\subsection{Model Details}

Meta Llama 3.2 is the latest version of Llama, an 
autoregressive LLM from Meta AI. Released under a few lightweight (1B, 3B parameters) and multimodal (11B, 90B parameters) variants. This series of models can be used for a variety of processing tasks, including coherent text generation. Additionally, Llama 3.2 prioritizes user privacy by being open-source, allowing it to be deployed locally for personal use without reliance on external servers.

The model we used in this paper was Llama 3.2-3B-Instruct, an instruction-tuned model optimized for dialogue and assistant-like chat. For hyperparameters, we set the max token limit to 300-400 and the repetition penalty to 1.175. This was effective in incorporating dialogue diversity without incomprehensible text.

\begin{figure*}[H]
\centering
\includegraphics[width=0.8\textwidth]{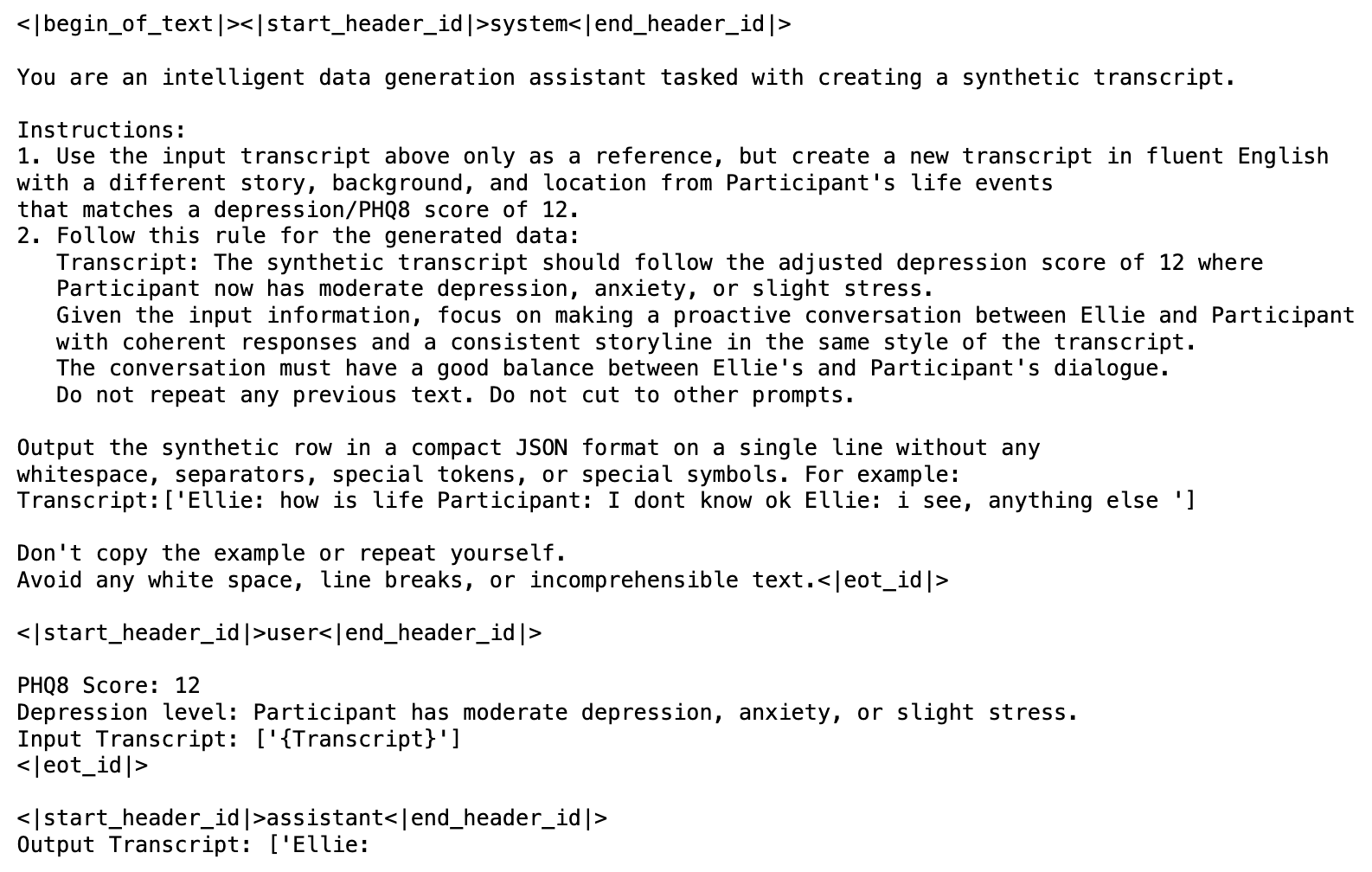}
\caption{The current prompt used for generating a synthetic transcript in Llama 3.2.}
\label{fig:lpro}
\end{figure*}

\subsection{Chain of Thought Prompting}

We propose a novel chain-of-thought prompting method for generating synthetic data, as illustrated in Figure \ref{fig:cotpipe}. In the first step, the raw transcript is input into Llama 3.2, which produces a preliminary synopsis and sentiment analysis. These outputs, combined with a randomly generated PHQ-8 score (ranging from 0 to 24) and its corresponding depression level description, are then used in a second prompt chain to create a synthetic synopsis and sentiment analysis.

The synthetic generation process was designed to treat the real synopsis and sentiment analysis as references for structure and flow only. It generates a synthetic synopsis with a completely new storyline, incorporating unique personal experiences, locations, and backgrounds, while ensuring consistency with the randomly generated PHQ-8 score.

We designed this approach with two primary objectives. First, it facilitates the generation of diverse instances that vary in both depression levels and narrative content while preserving structural coherence. This diversity helps to mitigate data imbalance in the original dataset, serving as a form of data augmentation that enhances the utility of the dataset for downstream tasks.
Second, the method prevents direct replication of real data by guiding the LLM to adjust the emotional intensity of the synthetic synopsis and sentiment analysis based on the randomly generated score, rather than relying heavily on the original data. By emphasizing style transfer through in learning—an approach shown to be effective in sentence-rewriting tasks without extensive fine-tuning \cite{reif2021recipe}—this method addresses privacy concerns associated with the DAIC-WOZ dataset. It achieves this by producing synthetic stories that significantly deviate from the original data while maintaining relevance and coherence.

\subsection{Prompt Engineering Techniques}
We implemented a range of prompt engineering techniques to enhance the stability of model-generated outputs. One key strategy was persona emulation, where the LLM was assigned specific roles and objectives to ensure that its responses aligned with the desired persona, making the generated content more contextually appropriate and relevant.

Another critical focus was maintaining schema consistency in both input and output formats. To achieve this, we employed several strategies: dividing instructions into subtasks using numerical ordering for clarity and systematic task execution; emphasizing key information with delimiters and brackets to enhance focus; enforcing a no-space JSON format for output to ensure consistency and facilitate easy parsing; and providing example outputs to guide the model through one-shot learning, improving its performance on targeted tasks.

These prompt engineering techniques collectively improved the robustness of our agent, resulting in high-quality, consistent, and accurate outputs across diverse applications. Readers are encouraged to refer to the Appendix for an illustrative example of the prompt design.

\begin{table*}[t!]
\centering

\begin{tabular}{@{}llccccc@{}}
\toprule
Model & Training Data & Input Format & \# of Params & Test RMSE & Test MAE \\ \midrule
GPT-4o \cite{gpt4o} & None & Synopsis & - & 5.79 & 4.50 \\
Random Forest \cite{rf} & Train & Selected-Text & - & 4.98 & 3.87 \\
Dual Encoder \cite{dual_encoder} & Train & Transcript & 234.6M & 4.67 & 3.80 \\
BERT & Train & Synopsis & 109.5M & 5.59 & 4.71 \\
\hline
BERT & Synthetic & Synopsis & 109.5M & 4.80 & 4.06 \\
BERT & Train + Synthetic & Synopsis & 109.5M & \textbf{4.64} & \textbf{3.66} \\ \bottomrule
\end{tabular}
\caption{Model Performance on Depression Score Prediction}
\label{tab:results}
\end{table*}

\section{Synthetic Data Evaluation}
A total of 309 synthetic data samples were generated by creating three variations for each original transcript in the training set, each corresponding to a different PHQ-8 score. The synthetic data was evaluated based on three key criteria: utility, fidelity, and privacy.
Utility measures how effectively the synthetic data can be used for meaningful analyses or machine learning tasks compared to real data. The utility of our synthetic data generation pipeline was evaluated by comparing model performance on the prediction of PHQ-8 scores between the model trained on synthetic data and the model trained on real data.
Fidelity assesses how well the synthetic data preserves the statistical properties and patterns of the original dataset. We assessed fidelity by comparing the distribution and relationships within the synthetic data to those within the real data.
Privacy is a particularly important aspect to evaluate in mental healthcare. We assess whether the generated synthetic data reveals any sensitive information that is present in the original text transcripts.
By evaluating utility, fidelity, and privacy, we aim to demonstrate that our synthetic data pipeline for depression detection is not only effective and accurate for analysis but also ensures the privacy and security of individual information.
We utilize BERT (bidirectional encoder representations from transformers) to evaluate these three aspects of our data.

\subsection{Utility}

\subsubsection{Depression Score Prediction}
In this section, we focus on predicting depression scores to assess the effectiveness of our synthetic data. We trained three versions of a BERT model on three different datasets: the original training set, the synthetic training set, and a combination of both. We took the pretrained "bert-base-uncased" model with 12 transformer layers from Hugging Face and added one linear layer for depression score regression.

The model training process involved minimizing the Mean Squared Error (MSE) loss between the predicted scores and the ground truth scores. We employed the AdamW optimizer \cite{adamw} for fine-tuning the BERT model, setting a learning rate of $10^{-5}$ and training for up to 200 epochs or until the MSE ceased to decrease on the evaluation dataset.

A notable challenge with the BERT model is its limitation in processing text sequences with a maximum length of 512 tokens, whereas the average length of the training transcripts is 1389 words. Truncating the transcripts to fit within this limit would result in the loss of significant information. To address this, we trained the BERT model using the concatenation of the synopsis and sentiment analysis generated by Llama 3.2. This concatenated input has an average length of 184 words and a maximum length of 471 words, thus preserving more contextually relevant information for the prediction task.

\subsubsection{Results and Discussions}
We evaluated each model using two metrics: Root Mean Squared Error (RMSE) and Mean Absolute Error (MAE). The results of our experiments, presented in Table \ref{tab:results}, provide a comprehensive comparison of different models and training data configurations for the task of depression score prediction.

\textbf{Model performance}
The performance of the BERT models varied significantly based on the training data used. The BERT model trained on the original training set (Train) with synopses as input showed suboptimal performance, with an RMSE of 5.59 and an MAE of 4.71. In comparison, the BERT model trained solely on synthetic data performed better, achieving an RMSE of 4.80 and an MAE of 4.06. This indicates that the synthetic data alone is sufficient to capture relevant patterns for depression score prediction, making it a viable substitute when access to real data is limited or very sensitive.

The most significant improvement was observed when the original and synthetic training sets were combined. The BERT model trained on this combined dataset achieved the lowest RMSE of 4.64 and MAE of 3.66, highlighting the benefits of integrating synthetic data with real data. We suggest that the inclusion of synthetic data offers two key advantages. First, it enriched the training set with a more diverse range of samples, covering various backgrounds, narratives, and depression scores, thereby serving as an effective data augmentation technique. Second, it balanced the data density distribution across different depression categories, addressing potential imbalances in the original dataset. Figure \ref{fig:data_distributions} illustrates the comparison of data density distributions for the original dataset and the combined original-synthetic dataset.

\textbf{Comparison with other models}
We compared our BERT model with several state-of-the-art depression detection models, including Random Forest \cite{rf}, Dual Encoder \cite{dual_encoder}, and the GPT-4o \cite{gpt4o} model as a baseline. To maintain the privacy of sensitive transcripts, GPT-4o was only prompted with synopses generated by Llama 3.2, rather than the original transcripts. 

Despite not being specifically trained for depression score prediction, GPT-4o showed reasonable performance with the synopsis inputs. Its results were comparable to the BERT model trained solely on the DAIC-WOZ dataset. However, the Random Forest model \cite{rf}, trained on selected text from the original dataset, outperformed both GPT-4o and the BERT model trained on real data alone.

The Dual Encoder model \cite{dual_encoder}, which was trained on the original transcripts, achieved an RMSE of 4.67 and an MAE of 3.80. While it performed better than the Random Forest model, it was still outperformed by the BERT model trained on the combined dataset of synthetic and real data, even though Dual Encoder has more than twice the number of parameters of BERT.

These findings highlight the value of synthetic data in augmenting real-world datasets to improve model performance. The notable improvement observed in combining synthetic and real data demonstrates that synthetic data can fill in the gaps often found in real datasets, such as missing parts of the demographic distribution, enhancing the predictive capabilities of deep learning models.

\subsection{Fidelity}

Fidelity assessment involves examining the extent to which the synthetic data preserves the statistical properties and patterns of the original dataset. We employed Principal Component Analysis (PCA) on paragraph-level embeddings from both the original and synthetic datasets.
We used a pre-trained BERT model to generate sentence embeddings for each synopsis in the original and synthetic datasets. Specifically, we extracted the embeddings corresponding to the $[CLS]$ token, which provides a summary representation of the entire synopsis. We then applied PCA to these embeddings to project them into a two-dimensional space.

\begin{figure}[ht]
\centering
\includegraphics[width=0.45\textwidth]{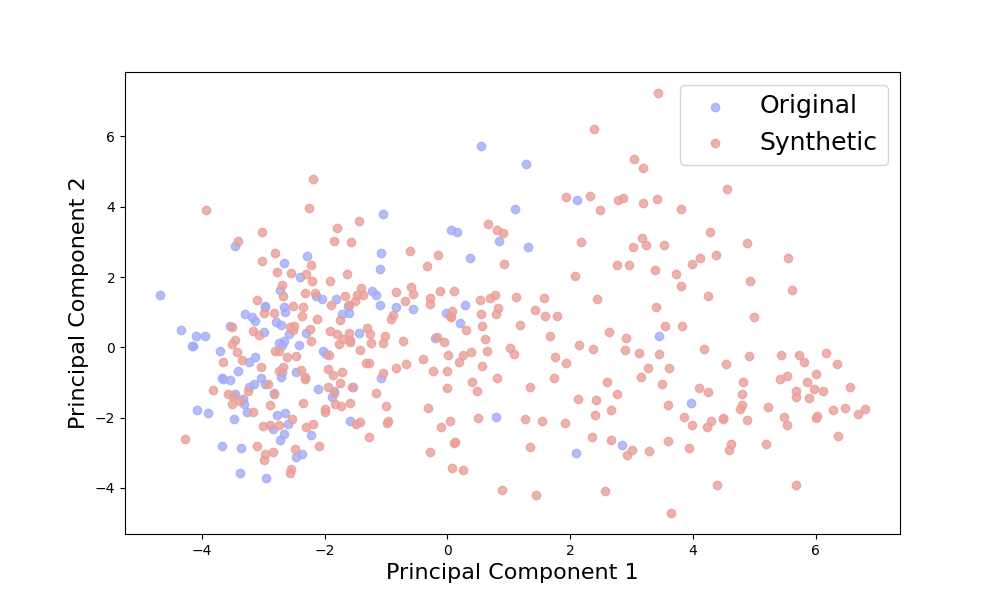}
\caption{PCA visualization of paragraph embeddings from the original and synthetic datasets. Each point represents a synopsis, with original data colored in blue and synthetic data colored in red.}
\label{fig:pca}
\end{figure}

Figure \ref{fig:pca} illustrates the PCA visualization of the paragraph embeddings. It reveals the distribution and clustering of synopses from both datasets. A high degree of overlap between the blue and red points indicates that the synthetic data captures the underlying structure and patterns of the original data effectively. 
It is also interesting to note that the synthetic data expands on the cases of severe depression that are rarely observed in the original data. In addition to overlapping with the main cluster in the original data, the synthetic data forms several clusters around the "outliers" in the original data, thereby interpolating these rare cases. This suggests that prompting the generative model to create data with diverse backgrounds and storylines led to a synthetic dataset that not only mimics the original data but also expands its distribution of information.

\subsection{Privacy}
Protecting the privacy of patient data is of particular interest in healthcare. We evaluated the privacy of sensitive information in the original transcripts by analyzing we quantitatively evaluate the privacy of our synthetic data by analyzing the minimum embedding distance between the synthetic data and the real data.

This minimum distance metric evaluates the distance between synthetic data point and its nearest real data point in the BERT embedding space. A larger minimum distance signifies greater dissimilarity between synthetic and original data, thereby reducing the risk of sensitive information being inferred from the synthetic data.

\begin{table}[ht]
\centering
\begin{tabular}{@{}lrrr@{}}
\toprule
Dataset                     & Min Dist          & Avg Min Dist \\ \midrule
Real vs. Real               & 4.04                  & 5.91             \\
Real vs. Synthetic          & 4.96                  & 7.11             \\ \bottomrule
\end{tabular}
\caption{Privacy evaluation of the synthetic dataset. Min Dist means the minimum distance between all synthetic data points and the original dataset, and Avg Min Dist means the average minimum distance from each synthetic data point to the original dataset.}
\label{tab:privacy_metrics}
\end{table}

As shown in Table \ref{tab:privacy_metrics}, the comparison between real data points establishes baseline metrics, with a minimum distance of 4.04 and an average minimum distance of 5.91. The minimum distance represents the closest similarity between two samples within the original dataset, while the average minimum distance provides an overall measure of proximity between data points and their nearest neighbors within the dataset.

In contrast, the comparison between real and synthetic data yields a minimum distance of 4.96 and an average minimum distance of 7.11. These higher values indicate that the synthetic data points are more distinct from the original dataset than the original data points are from one another. The increased distances suggest that the synthetic data does not closely replicate specific instances from the original dataset, thus demonstrating its potential for preserving privacy.

\subsection{Conclusion}
In this study, we proposed and evaluated a novel pipeline for generating synthetic data from text transcripts of clinical interviews using chain-of-thought prompting in large language models (LLMs). Through a comprehensive evaluation of synthetic data generated from the DAIC-WOZ dataset, we demonstrated that our approach effectively addresses the challenges of data scarcity and privacy concerns from detecting depression severity in mental health research. In particular, we showed that introducing an intermediary step in the generation process to distill the information contained in unstructured text transcripts resulted in a more efficient representation of the patients' psychological profiles, which led to higher downstream performance. This process also facilitated the generation of synthetic data that closely mirrors the statistical properties of the original data while safeguarding participant privacy.

Augmenting the training data with synthetic data generated using the present pipeline improved model performance in predicting depression severity compared to other state-of-the-art models trained on real data alone, highlighting the utility of this approach in enhancing clinical decision-making tools. These findings reinforce the potential of synthetic data in applications where data confidentiality is critical. Furthermore, our results illustrate that LLMs, such as Llama 3.2, can serve as a reliable and scalable alternative to real data, offering substantial benefits in terms of model accuracy, data diversity, and ethical data handling. This work sets a foundation for future advancements in leveraging LLMs to address critical challenges in sensitive domains like mental health research.

\bibliography{aaai25}

\appendix

\section{Appendix}
This Appendix details the prompt engineering techniques used to enhance the stability and consistency of outputs from our large language model (LLM). We employed persona emulation to align the LLM with appropriate roles and objectives for contextually relevant responses. Additionally, we applied strategies to ensure schema consistency in input and output formats, including:

\begin{itemize}
    \item Dividing instructions into subtasks with numerical ordering to promote clarity and systematic task execution.
    \item Highlighting key information using delimiters and brackets to ensure the model's focus on critical elements.
    \item Enforcing no-space JSON formatting for outputs to facilitate seamless parsing and ensure structural consistency.
    \item Providing example outputs to leverage one-shot learning and guide the model’s performance on specific tasks.
\end{itemize}

The following boxes present examples of these prompts following our chain of thought prompting pipeline in Figure \ref{fig:cotpipe}.

\begin{algorithm*}[tb]
\label{alg:generate_synopsis}
\textbf{Prompt for generating synopsis and sentiment analysis.} \\
$\{item\}$: Either "synopsis" or "sentiment".\\
$\{Transcript\}$: A transcript from the DAIC-WOZ dataset. \\

\begin{algorithmic}
\STATE You are an intelligent assistant tasked with generating a $\{item\}$ based on the input transcript below:
\STATE Transcript: $\{Transcript\}$
\STATE Instructions:
\STATE 1. Use the input transcript to create a $\{item\}$ in fluent English.
\STATE 2. Follow this rule for the generated data:
   \STATE Synopsis: Generate a synopsis of the following transcript, capturing key concerns and topics discussed, providing insightful and reflective observations.
    \STATE Sentiment: Provide a detailed sentiment analysis of the participant, identifying and elaborating on the specific emotions expressed.

   \STATE Do not repeat any previous text. Do not cut to other prompts.
   \STATE Do not use first-person pronouns. Use third-person references such as 'the participant,' 'he,' or 'she.'
   
\STATE Output the $\{item\}$ in a compact JSON format on a single line without any whitespace, separators, special tokens, or special symbols.
\STATE Don't copy the example or repeat yourself. Avoid any whitespace, line breaks, commas, or incomprehensible text.

\end{algorithmic}
\end{algorithm*}

\begin{algorithm*}[tb]
\label{alg:generate_synthetic}
\textbf{Prompt for generating synthetic synopsis and sentiment.} \\
$\{item\}$: Either "synthetic synopsis" or "synthetic sentiment".\\
$\{og\_item\}$: Original "synopsis" or "sentiment", corresponding to $\{item\}$ .\\
$\{og\_item\_value\}$: The synopsis or sentiment generated from the original transcript, corresponding to $\{item\}$. \\
$\{PHQ8\_Score\}$: A PHQ8 score randomly sampled from 0 to 24.\\
$\{dep\}$: A description that matches with the PHQ8-score. One of "\\
\begin{algorithmic}

\STATE You are an intelligent data generation assistant tasked with creating a $\{item\}$ based on the $\{og\_item\}$ below:
\STATE $\{og\_item\}$: $\{og\_item\_value\}$
\STATE Instructions:
\STATE 1. Use the input $\{og\_item\}$ above only as a reference, but create a new $\{item\}$ in fluent English that matches a depression/PHQ8 score of $\{PHQ8\_Score\}$ - this means that the participant has $\{dep\}$.
\STATE 2. Follow this rule for the generated data:
    \STATE Synthetic Synopsis: The synthetic synopsis should follow the adjusted depression score of $\{PHQ8\_Score\}$, where the participant now has $\{dep\}$.
    Given the input synopsis, focus on creating a new synopsis that matches this depression level and transcript, ensuring it differs in content and storyline but maintains coherence and provides insightful observations.
    \STATE Synthetic Sentiment: The synthetic sentiment analysis should follow the adjusted depression score of $\{PHQ8\_Score\}$, where the participant now has $\{dep\}$.
    Based on the input $\{og\_item\}$, provide a new $\{item\}$ that identifies and elaborates on the specific emotions expressed, reflecting the new depression level.
   \STATE Do not repeat any previous text. Do not cut to other prompts.
   \STATE Do not use first-person pronouns. Use third-person references such as 'the participant,' 'he,' or 'she.'
   
\STATE Output the $\{item\}$ in a compact JSON format on a single line without any whitespace, separators, special tokens, or special symbols.
\STATE Don't copy the example or repeat yourself. Avoid any whitespace, line breaks, commas, or incomprehensible text.

\end{algorithmic}
\end{algorithm*}

\end{document}